# Semantic Parsing for Question Answering over Knowledge Graphs

Sijia Wei[1], Wenwen Zhang[1], Qisong Li[2], Jiang Zhao[1]
1. School of Computer Science, Jianghan University
2. School of Computer Science, Wuhan University of Technology

**Abstract**

In this paper, we propose a novel method for question answering over knowledge graphs based on graph-to-segment mapping, designed to improve the understanding of natural language questions. Our approach is grounded in semantic parsing, a key technique for interpreting question utterances. The main challenges arise from handling implicit entities and relations, as well as complex constraints such as temporal conditions, ordinality, and aggregation within the context of a knowledge graph. To address these issues, our framework integrates both rule-based and neural methods to parse and construct accurate, comprehensive semantic segment sequences. These sequences are then assembled into semantic query graphs, providing precise representations of question utterances. We formulate question semantic parsing as a sequence generation task, employing an encoder-decoder neural network to map natural language questions into semantic segments. Furthermore, to enhance the identification of implicit entities and relations, we incorporate a graph neural network that leverages knowledge graph context to enrich question representations. Experimental evaluations on two benchmark datasets demonstrate the effectiveness and superior performance of our model in semantic parsing for knowledge graph question answering.

## 1 Introduction

The goal of Question Answering over Knowledge Graphs (KGQA) is to leverage the facts stored in a knowledge graph to answer natural language questions. The core challenge lies in understanding question semantics and aligning them with knowledge semantics. In recent years, deep neural networks have become the dominant approach for advancing KGQA, with research methods primarily divided into ranking-based and translation-based approaches [1].

The ranking-based approach learns vector representations of questions and candidate answers, using end-to-end models to score and rank candidates according to contextual information in order to produce the final answer. Early work [2] pioneered the use of neural networks for KGQA by matching questions with subgraph embeddings for scoring. Subsequent studies [3–5] enriched question representation with additional contextual information to reduce the search space, while others [6–8] enhanced the network architecture with attention mechanisms, multilinear convolution, or bidirectional RNNs to improve ranking accuracy. Another line of work [9–12] focused on scoring and ranking candidate logical forms corresponding to the question. While these methods remove the need for manual rules and vocabularies, they suffer from limitations such as insufficient semantic modeling, reliance on central entities for candidate generation, limited interpretability, and the inability to support user interaction in the intermediate process.

Translation-based methods, on the other hand, parse question semantics into machine-interpretable logical forms, which are then executed or reasoned over the knowledge graph to derive answers. SPARQL remains the most common logical form in KGQA [13–15], though alternatives such as Lambda-DCS [16,17], CCG [18,19], and FunQL [20] have also been proposed to enhance semantic representation. Traditional generation algorithms [1,19] rely heavily on predefined grammars and dictionaries, which are complex and difficult to adapt across domains. By contrast, deep learning–based semantic parsing methods are inspired by translation models, framing the parsing process as a Seq2Seq problem [21–23]. These approaches employ encoder–decoder architectures to generate logical forms, leveraging the strong sequence prediction capabilities of recurrent neural networks. However, treating questions purely as sequences makes it difficult to capture structural information, while also overlooking textual information linked to the knowledge base.

A knowledge graph typically contains information about entity types, relationship categories, and their associated instances. The relationships, represented as subject–predicate–object (SPO) triples, can be broadly divided into two categories: those that connect entities to other entities, and those that link entities to their textual attributes. A formal representation of a domain knowledge graph is: $G = (T, R, E, I)$ , where $T = \{t_1 t_2, \cdots t_{|T|}\}$ is a set of entity types $E = \{e_1 e_2, \cdots e_{|E|}\}$ is a set of entities in which the type of the entity is one of the elements of $T$ . $R = \{r_1 \quad r_2, \cdots, r_{|R|}\} r_i \in R^e \cup R^l$ is a set of binary relation types, where $R^e$ denotes the relation between an entity and an entity $R^l$ denotes the relation between an entity and its literal attributes. $I = R^e \cup R^l$ is the set of relationship instances , whose elements are denoted as $r(e_1, e_2)$, where $r \in R$, $e_1 e_2 \in E$ . As illustrated in the example knowledge graph in [15], domain-specific Knowledge Graph Question Answering (KGQA) is generally centered on domain-related queries. The process of understanding these questions can be divided into three subtasks [1,10]: entity linking, relation identification, and constraint identification involving logical and numerical operations. Beyond the challenges of handling multiple entities and relationships [24] and incorporating temporal information [25–27], KGQA also faces several additional challenges.

Domain-specific knowledge graph question answering (KGQA) faces several challenges. One is the presence of implicit entities and relationships. Domain questions often omit explicit references to entities or relations, assuming that they are clear from the domain context. For example, in the question "How many states are there?" the intended meaning defaults to U.S. states based on the knowledge graph context. Similarly, in "How many rivers does Alaska have?" the relationship between rivers and Alaska cannot be fully understood from sentence semantics alone and requires contextual knowledge from the graph. Existing semantic analysis methods do not fully exploit such domain knowledge [28]. Another challenge is the handling of constraints, such as temporal conditions, ordering, and aggregation, which make semantic interpretation more complex [11]. Current intermediate representations or logical forms are primarily designed for general domains [20,28] and do not adequately capture domain-specific features, such as large-granularity reuse of fragments. A further difficulty is the combination of multiple intents within a single question. While simple questions usually involve a single intent, questions in domain-specific KGQA often contain multiple intertwined intents. Achieving accurate semantic parsing in these cases requires effectively decomposing and recombining intents [29,30].

To address these challenges, we propose SPEDN (Semantic Parsing with graph-to-sEgment mapping for Domain kNowledge graphs), a semantic parsing framework for intent understanding in domain KGQA. SPEDN introduces six semantic chunking patterns designed to represent the intents of domain-specific questions, and performs semantic parsing through the generation and assembly of semantic chunks. Complex question semantics are decomposed into a sequence of semantic chunks, each representing a minimal semantic unit and corresponding to a single-step query or inference on the knowledge graph. For example, the question “How many capitals does Rhode Island have?” can be decomposed into three semantic chunks: (1) the count of capitals, (2) a capital located in a given state, and (3) the state identified as Rhode Island. Each chunk represents part of the question intent, and the complete intent can be reconstructed by assembling the sequence. In this case, the state in chunk (2) is replaced by the entity set in chunk (3), which is then combined with chunk (1) to yield the full semantic representation of the question.

The main contributions of this paper are as follows. First, we propose SPEDN, a semantic parsing framework for KGQA that defines semantic chunking patterns to represent question semantics. Second, we model question parsing as the generation of semantic chunk sequences, using graph neural networks to parse questions into such sequences. Experiments on two datasets validate the effectiveness of this approach. Third, to address the challenge of implicit entities and relations, we incorporate contextual information from the knowledge graph by constructing a contextual dictionary of the question, thereby improving the performance of semantic parsing.

## 2 Question Semantic Parsing

The architecture of our approach is shown in Fig. 1. Given an input natural language question, a context dictionary is first generated with the aid of the knowledge graph, which provides additional information to the graph neural network. The network output is then used to generate semantic chunks that represent the intent of the question. These chunks are subsequently assembled into a semantic query graph, which serves as the parsed representation of the question intent. The following sections describe the components of the framework in detail.

### 2.1 Semantic Query Graph

Following the approach in [5,10,11,25], we represent questions as semantic query graphs. For example, Fig. 1(d) illustrates a semantic query graph for an interrogative sentence, where circles denote answers, rounded rectangles indicate entity types in the knowledge graph, solid arrows represent relationships between entities, and diamonds correspond to operations such as constraints or aggregation applied to entities. Given a question, the construction of its semantic query graph can be viewed as a process of sequentially identifying semantic chunks and incrementally assembling them into the graph. The dashed rectangle in the figure highlights how semantic chunks are combined to form the complete semantic representation of a question.

We define the following semantic block schema to represent the intent of the problem (where $p_t$ denotes a schema of type $t$, and $p, t$ is the type of element in the set of entities represented by the semantic block of this schema). An entity schema represents a collection of entities of a specified type. For example, entity(city) represents all cities, entity(country, id, 'us') represents the country with identifier “us,” and entity(flight, month, 'july') represents flights occurring in July.

$$\text{entity}_t = \{(t,\text{atrr},\text{value}): t \in T, \text{attr} \in \mathbf{R}^l\} \quad (1)$$

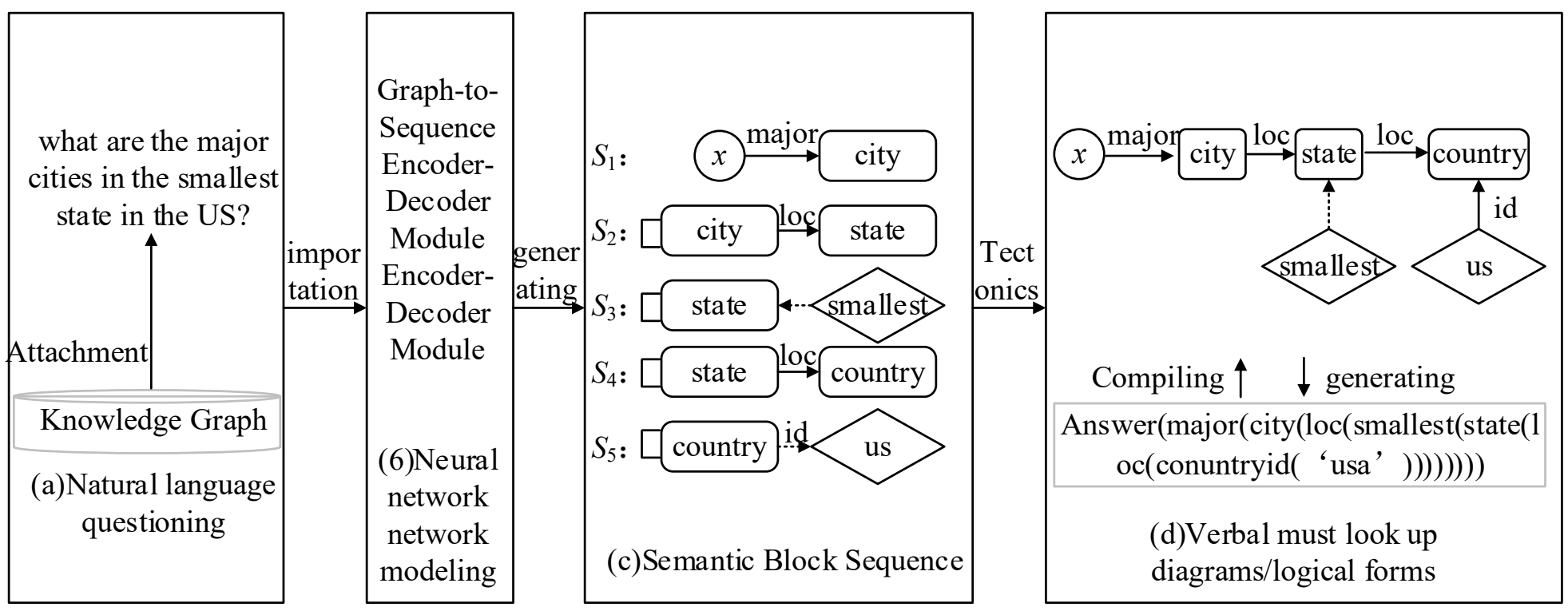

Fig. 1 Schematic diagram of the semantic parsing process based on semantic chunks for the question.

Relational schema: denotes the set of left entities of type $t$ obtained by reasoning from the set of right entities of type $t_2$ through the relation $r$, e.g. relation ( state, loc,: country) denotes the state etc. that has a loc relation with the specified country.

$$relation_{\,t} = \{(tp\{e: e \in t_2\}): p \in \mathbf{R}^e, (tt_2) \in T\} \quad (2)$$

Attribute schema: represents the result of a computation on an entity attribute whose object is of type literal in an SPO triple, e.g. literal(len, :river):

$$literal_{\,t} = \{(p, \{e: e \in t\}): p \in \mathbf{R}^l\ t \in T\} \qquad (3)$$

Sorting mode: indicates that after sorting, we get the entity at the specified position, such as ordinal (max,: state), ordinal (min,: city) and so on:

$$ordinal_{\,t} = \{(\mathrm{ord}\,\{e: e \in t\}): \mathrm{ord} \in \{\max, \min\}\ t \in T\}(4)$$

Aggregate mode: represents a single value after counting the set of entities specified by the input, e.g. aggr (count, :city):

$$\mathrm{aggr}_{n,n\in R} = \{(\mathrm{aggr}, \{e: e \in t\}): \mathrm{aggr} \in \{\mathrm{count\ average}\}\ t \in T\} \qquad (5)$$

Set Mode: represents the set of entities after computing the specified operations of the set such as intersection and union for the input set of two or more entities:

$$\mathrm{join}_t = \{(\mathrm{join}, \{e: e \in t\}\ \{e: e \in t\}): \mathrm{join} \in \{\ \mathrm{intersetion, union, exclude}\ \}\ t \in T\} \qquad (6)$$

Examples and illustrations of the various patterns are listed in Table 1.

2.3 Question Semantic Parsing

Our semantic parsing framework learns semantic parsing models based on the input natural language interrogative sentences and their corresponding semantic query graphs on the domain knowledge graph to accomplish question intent parsing. As shown in Fig. 1, given an interrogative sentence $X$, the context dictionary $X^C$ of the interrogative sentence is generated with the help of knowledge graph $G$ and inputted into the pre-trained graph neural network model to get the corresponding semantic block sequence $Y$ which is the semantic query graph corresponding to the interrogative sentence. The use of $Y$ can be directly matched on the knowledge graph to get the answer 1, or can be transformed into the corresponding logical form of reasoning to get the final answer 1. Therefore, the semantic parsing of a question is mainly divided into two steps: the construction of the question context dictionary and the generation of semantic block sequences.

Problem context dictionary construction: using $X$ and $G$ one can get the category $T^{\mathrm{in}}$ of candidate entities associated with $X$, and based on $G$ one can get the relationship $R^{\mathrm{in}}$ between elements in $T^{\mathrm{in}}$, thus obtaining the problem representation $X^c = \{X, T^{\mathrm{in}}, R^{\mathrm{in}}\}$ with contextual information as an input to the SPEDN model, and we will detail the processing in Section 3.1. Sequence generation of semantic chunks: using an encoder-decoder network, an input problem $X^C$ can be parsed into a sequence of semantic chunks $Y$. We need ① a graph encoder, which encodes the input $X^C$ into a vector representation, and ② a decoder, which is used to generate $Y$ conditional on encoding vectors, and we elaborate on the network model in Section 3.

## 3 Semantic Block Generation Neural Network

We use an encoder-decoder based graph neural network to accomplish the generation of semantic block sequences based on the network architecture in [31]. As in Fig. 3, the model mainly consists of a graph encoder consisting of a node embedding layer and a graph embedding layer, a node attention mechanism, and an RNN-based sequence decoder, and the network model is described in detail below.

3.1 Question Preprocessing

Semantic Normalization: We first employ WordNet to convert all words in the input sentence X into their canonical forms. This step unifies different morphological variants (e.g., tense, plural forms) into a single representation. For instance, “cities” is normalized to “citi” by WordNet.

Candidate Set Construction: Using the input knowledge graph Gin, we identify entities directly connected to those mentioned in the question and collect them into a candidate set. The process is as follows:

(1) Entity Expansion: For each entity in the input, we retrieve all adjacent entities from the knowledge graph and add them to the candidate set Cin. Each entity is associated with a relation rin that denotes the type of connection.

(2) Relation Expansion: We further retrieve all relations directly connected to the input entities in the knowledge graph and include them in the candidate set Rin.

Candidate Relationships: Based on the candidate entities and relations, we construct the possible relationships among them. At the same time, we establish links between the entities and relations to form a coherent candidate structure.

Question Graph Construction: Finally, we transform the input question into a graph representation by connecting the input entities with their associated candidate entities and relations. These are integrated into Rin, resulting in a preliminary question graph that serves as the input to the neural network.

3.2 Graph Encoder

Node Vectors: Nodes are divided into two categories. The first category corresponds to the words appearing in the input question, forming a dictionary $V'$. We initialize the vectors in $V'$ randomly. The second category corresponds to candidate entity types connected to the question, forming a dictionary $V''$. This dictionary is constructed from entity categories in the knowledge base $T$, and vectors are initialized randomly.

Edge Vectors: After question embedding, edges are divided into two categories. The first category contains labeled edges, where each label corresponds to a relation type in the knowledge base $R$. The second category consists of unlabeled edges, which we initialize randomly using a shared vector for all unlabeled edges.

Node Embedding Layer: Following [31], node embeddings are generated by aggregating information from neighbors. For each node $h^i$, the representation is formed by concatenating information from both in-neighbors and out-neighbors:

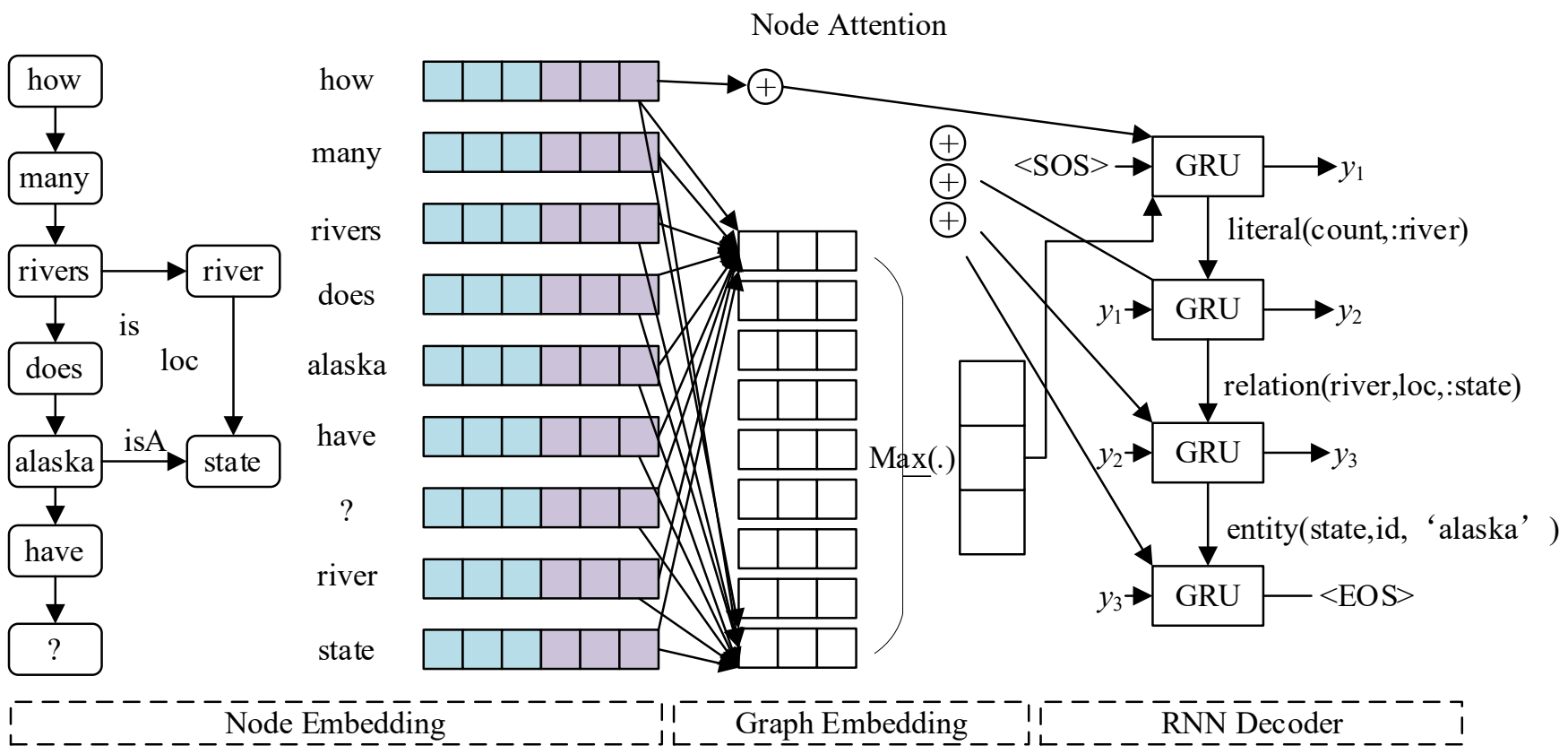

Figure 3 SPEDN framework neural network architecture

$$h_{i'} = [h_i^{in}, h_i^{out}] \quad (7)$$

Graph Embedding Layer: A pooling operation is applied over node embeddings to obtain a representation for the entire question graph. Specifically, we use a max-pooling strategy.

3.3 Sequence Decoder

We employ a sequence decoder [30] to generate a semantic block sequence. The decoder is a bidirectional GRU model, where each decoding step generates a semantic block, ultimately forming the sequence of semantic blocks. At each step $i$, the decoder updates its hidden state $h_i^d$ based on the previously generated block, the current state, and the predicted output $s_i$:

$$s_i = \tanh\left(W^d [v_i^n, h_{i-1}^d]\right), \quad (8)$$

$$s_{i+1} = \mathrm{GRU}\left(\left[\phi^{(v)}(y_i), c_i\right], s_i\right) \quad (9)$$

Semantic Block Embedding: Each semantic block is represented by two components: name and parameters. The name specifies the type of the semantic block (e.g., a structural unit), while the parameters represent semantic details. For instance, the block `relation(city, loc, state)` has the name relation(city) and parameters loc and state. By separating structure and semantics, we allow parameter embeddings to be shared across different block types.

Beyond blocks of type `entity` and `relation`, other block types such as `class` and `attribute` are also included. We uniformly represent both structure and semantic details with randomly initialized embeddings.

Semantic Decoding Control: At each step, the decoder predicts three elements of a semantic block: block name, entity type, and parameters. Beam search is applied to generate candidate sequences, and parameter embeddings are matched with predicted block names and entity types. The decoder then outputs the most probable semantic block sequence.

3.4 Node Attention Mechanism

The sequence decoder is modeled as an RNN, which uses the predicted result $\gamma_i$, the hidden state of the RNN at this step, and the contextual representation $c_i$ to generate the $i$-th output $y_i$. The attention representation $c_i$ at the decoding stage corresponds to the network's attention over the node representations in the question graph. Each node representation encodes information related to the input question graph while simultaneously attending to neighboring nodes centered around the current node.

The attention-weighted representation $c_i$ is computed as follows:

$$c_i = \sum_{j=1}^{|V'|} a_{ij} h_j, \quad \text{where} \quad a_{ij} = \frac{\exp(e_{ij})}{\sum_{k=1}^{|V'|} \exp(e_{ik})}, \quad e_{ij} = a(s_{i-1}, h_j) \quad (10)$$

Here, $a_{ij}$ denotes the attention weight for each node representation, which is controlled by the similarity between the current hidden state $s_{i-1}$ of the RNN and the representation $h_j$ of node $j$ in the question graph. The attention mechanism evaluates the compatibility between the decoder's current hidden state and the node representations of the input graph.

We employ a feed-forward neural network parameterized by $a$ to compute this compatibility score. This neural network is jointly trained with the other components of the G2S model.

4 Experiments

This section discusses the experiment and compares it with the existing research to verify the performance of the method in this paper.

4.1 Dataset

We verify the performance of SPEDN model in two data sets. GE0880 is a question-and-answer data set about American geographic information, which contains 880 question-and-answer pairs. The questions are described in natural language, and the answers are in logical form in Prolog format. Referring to reference [18], we use the standard 600/280 to establish training sets and test sets. ATIS is a question-and-answer data set about civil aviation flights, which contains nearly 5,000 natural language question-and-answer pairs about flights, and the logical form is Lambda-DCS. We use standard 4473/448 to build training sets and test sets. In order to train the semantic block sequence neural network based on the network architecture in 3, we preprocess the logical forms of the answers of the training and test data in the two data sets and convert them into the form of semantic block sequences, as shown in Table 2.

4.2 Experimental setup

Our model includes parameters of graph encoder, parameters of GRU $W^{(s)}$, word embedding in question $\phi(x)$ and semantic block embedding in model sequence $\phi(y)$. Using supervised method, the question $X$ and its corresponding semantic block sequence $Y$ in the training corpus are obtained by training the training corpus with these parameters. We use maximal function to generate $Y$ corresponding to $X$, and the objective function is: $Y^* = \sum_{i=1}^{n} \log P(Y_i \mid X_{<i})$.

In reference [31], we use Adam optimizer to train and update parameters, the batch size is set to 30, and the learning rate is set to 0.01. In order to avoid over-fitting, we apply dropout strategy [36] in the decoding layer, and set the ratio to 0.2. In the graph encoder, the default jump point k size is set to 3, the initial feature vector size of the node is set to 100, and ReLU is used as a nonlinear function, and the parameters of the aggregator are randomly initialized. The decoder has one layer and the hidden state size is 256. Since SPEDN with mean aggregator and pool-based graph embedding usually performs better than other configurations, we use this setting as the default model. The Beam size is set to 5 when the decoder outputs. We train the model by iterating 80 times. The model is realized by Pytorch, and $Y^*$ is predicted by using test set in each iteration, and the error between $Y^*$ and $Y$ is accumulated to the loss value.

4.3 Experimental results

In this section, we discuss the effect of SPEDN model on GEO and ATIS data sets. Based on the method described in this paper, we use three configurations to train the analytical model: the first is the basic SPEDN model; The second is to decompose semantic blocks and embed them into the decoding layer of the network to realize the prediction of semantic block sequence (SPEDN(+MP)); The third one adds a sequence parsing controller (SPEDN(+MP+Controller)) on the basis of the second one to control the legality of assembling semantic block sequences into query graphs. We use the accuracy on the test set to evaluate the performance of each system. Table 3 lists the comparison between our model and the experimental results of existing research.

The accuracy of the model on GEO is 86.4%, and it reaches 88.3% after adding semantic block decomposition and embedding. The decoding controller improves the prediction accuracy, reaching 90.5%. At the same time, the results of three configurations on ATIS data set are 83.9%, 85.1% and 85.7%, respectively. Combining the strong representation ability of semantic query graph and the strong prediction ability of encoder-decoder model, especially based on graph neural network, it can be combined with the prediction ability of encoder-decoder model.

Semantic block embedding: our model embeds each semantic block independently by default, and we also use the structure of semantic blocks and the way of semantic information decomposition and embedding to carry out experiments. The experimental results show that it has a great influence on the accuracy, which is improved by 1.9 on GEO data set and 1.2 on ATIS data set. It can be seen that decomposing large-grained semantic block information helps to share the contextual information of each part, and can effectively improve the prediction level of the model on small-scale data sets such as GEO and ATIS.

Table 2 Examples of question and answer pairs and their corresponding semantic blocks in GEO and ATIS data sets. The second column gives the average length of sequences in data sets, and it can be seen that semantic blocks can effectively reduce the length of sequences.

| data set | average length | example |
|---|---|---|
| GEO | 7.6 | interrogative sentence: what are the major cities in the smallest state in the us? |
| | 28.2 | Logical form: answer( A，( major( A) ， city( A) ， loc( A， B) ， smallest ( B，( state ( B) ， loc ( B ， C) ， const ( C， countryid( usa) ) ) ) ) ) |
| | 2.9 | Semantic block sequence: literal ( major， : city ) relation ( city， loc ， : state ) ordinal ( smallest， : state ) relation ( state， loc， : country) entity( country， id， ‘usa’) |
| AITS | 11.1 | Question: what are the flights between dallas and pittsburgh on july eight ? |
| | 28.4 | Logical form: ( _lambda $ 0e ( _and( _flight $ 0) ( _ from $ 0 dallas: _ci) ( _to $ 0 pittsburgh: _ci) ( _ day_number $ 0 8: _dn) ( _month $ 0 july: _mn) ) ) |
| | 6.1 | Semantic block sequence: entity ( flight) relation ( flight， from， : city) entity( city ， id ， ‘dallas’) relation( flight， to， : city) entity( city， id， ‘pittsburgh’) entity( flight ， day _number， ‘08’) entity( flight ， month， ‘july’) |

Table 3 Accuracy on GEO and ATIS datasets

| System | CEO | ATIS |
|---|---|---|
| Zettlemoyer and Collins ( 2007) [19] | 86. 1 | 84. 6 |
| Kwiatkowski et al. ( 2011) [37] | 88.6 | 82. 6 |
| Liang et al. ( 2011) [17] ( + lexicon) | 91.1 | — |
| Zhao et al. ( 2015) [38] | 88.9 | 84.2 |
| Rabinovich et al. ( 2017) [39] | 87.1 | 85.9 |
| Jia and Liang ( 2016) [21] ( + data) | 89.3 | 83.3 |
| Dong andLapata ( 2016) [23] : 2Seq | 84.6 | 84.2 |
| Dong andLapata ( 2016) [23] : 2Tree | 87.1 | 84. 6 |
| Seq2Act ( + C1 + C2) [28] | 88.9 | 85.5 |
| SPEDN | 86.4 | 83. 9 |
| SPEDN( + MP) | 88.3 | 85.1 |
| SPEDN( + MP + Controller) | 90.5 | 85.7 |

Decoding controller: Adding the decoding controller has different effects on the prediction accuracy of the two data sets of the model, which is improved by 2.5 on GEO data set and 0.7 on ATIS data set. From the experiment, it can be seen that the decoding controller performs optimal control based on entity types, while the number of entity types on ATIS data set is less, so the influence on the results is smaller than that on GEO data set. At the same time, the controller can ensure that the generated semantic blocks can be correctly assembled into a semantic query graph.

## 4.4 Experimental analysis

Semantic block length: Table 2 shows the average length of logical forms and semantic blocks in each data set. It can be seen that comparing the sequence length of actions in Seq2Act 2 (GEO is 18.2, ATIS is 25.8) with the sequence length of original logical forms of data sets (GEO is 28.2, ATIS is 28.4), Semantic block can effectively reduce the length of the predicted target sequence. The length of SPEDN semantic block sequence is 10.3% of the logical form of GEO data set, 15.9% of that of Seq2Act on GEO, 21.5% of that of ATIS data set, and 23.6% of that of Seq2Act on ATIS. The decrease of sequence length has a great influence on improving the accuracy of prediction and effectively reducing the complexity of semantic map construction.

Semantic block distribution: The transformed semantic block pattern and entity distribution are shown in Figures 4(a) and 4(b). As can be seen from the figure, the semantic block lengths in GEO data set are mainly distributed in 2 and 3, accounting for 76.9% of all data, while those in ATIS data set are mainly distributed in 4~6, accounting for 78.6% of all data. From the distribution of semantic block patterns in Figure 4(c), entity patterns and relational patterns account for most of the semantic block patterns of the two data, with GEO being 64. ATIS reached 82.3%. In addition, attribute mode and sorting mode are more common in GEO data set (31.4%), while set mode is the main mode in ATIS data set (15.7%). The data distribution of the mode reflects the characteristics of problems in different fields.

Graph neural network: By embedding the problem into the neural network graphically, more contextual information can be added, which greatly improves the effect of semantic analysis of the model. We use two ways of graph construction: chain connection and full connection chain connection, which only connect the word in the question with the word after it, and full connection, which connects each word in the question with all the words after it. Experiments show that the two ways have little influence on the experimental results.

## 4.5 Error Analysis

We have analyzed the output error information, and the semantic block sequence parsing errors are mainly divided into the following two categories.

Loss of problem semantics: The attention model does not consider the alignment history, so that some words are ignored in the parsing process. For example, in the first and second cases in Table 4, "which border texas" and "big" will be ignored in the decoding process, so that the further definition of the problem intention is missing in the parsed result. This problem can be further solved by using the explicit word coverage model used in neural machine translation.

Semantic understanding error: Because the data of GEO and ATIS are relatively small, there will be words or questioning methods in the test set that have never been seen in the training set, which makes the prediction results of the model biased. In the third case in Table 4, the correct result can be obtained by using "how many" to ask questions. One solution is to learn word embedding on the annotated text data, and then use it as a pre-training vector for the problem words, or change it into a form that can be recognized by the model [41-43].

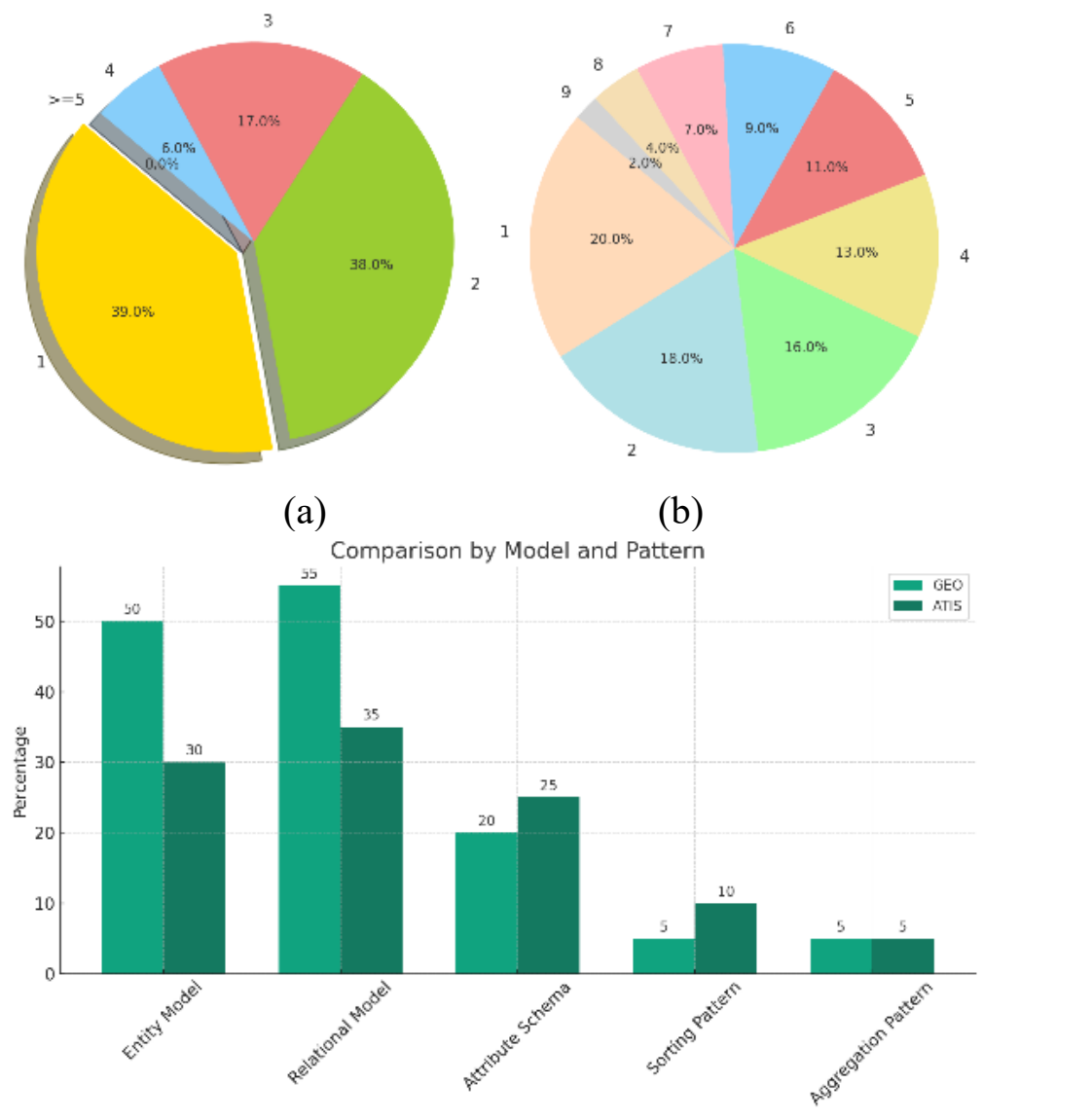

(c)

Fig. 4 The distribution of sequence length after the logical forms of answers in GEO (a) and ATIS (b) datasets are transformed into semantic blocks, (c) the distribution of semantic block patterns in two data sets.

Table 4 Some error analysis examples, each of which includes questions, the best answers and the results predicted by the model.

| type of error | example |
|---|---|
| Missing question Topic semantics | Question: what are the populations of states which border texas? Answer semantic block: literal( population，: state) relation( state，next_to，: state) entity( state，id，'texas') Predictive semantic block: literal( population，: state) entity( state，id，'texas') |
| Missing question Topic semantics | Question: how many big cities are in pennsylvania ? Answer semantic block: aggr( count，: city) join( intersection，: city，: city) entity( city，major，1 ) relation ( city，loc，: state ) entity ( state，id，'pennsylvania') Predictive semantic block: aggr( count，: city) relation( city，loc，: state) entity( state，id，'pennsylvania') |
| Semantic principle Solve mistakes | Question: give me the number of rivers in California Answer semantic block: aggr( count，: river) relation( river，loc，: state) entity( state，id，'california') Predictive semantic block: literal( len，: river) relation( river，loc，: state) entity( state，id，'california') |

## 5 Conclusion

In this paper, we propose SPEDN, a semantic analysis framework for knowledge graph question answering that integrates the accuracy of rule-based methods with the coverage of deep learning, while also incorporating contextual information from the knowledge graph. The semantic analysis task is modeled as a graph-to-sequence encoder–parser problem, where the analysis of semantic blocks is independent of the logical form of the QA output, ensuring strong adaptability. We evaluate the framework on two datasets, and the experimental results demonstrate its effectiveness.

To further enhance the performance of the semantic parsing framework, we plan to design a semantic block combination algorithm that improves the reusability of coarse-grained semantic parsing. Additionally, to address the challenge of limited training data, we will develop an interactive automatic or semi-automatic corpus generation algorithm and tool for domain-specific knowledge graphs, building on related research. Through an interactive user interface, this tool will enable rapid construction of QA corpora for specific domains without requiring extensive domain expertise.